\documentclass[10pt,twocolumn,letterpaper]{article}

\usepackage{wacv}
\usepackage{times}
\usepackage{epsfig}
\usepackage{graphicx}
\usepackage{amsmath}
\usepackage{amssymb}

\newcommand{\figname}{Figure~}
\newcommand{\tabname}{Table}

\usepackage[pagebackref=true,breaklinks=true,letterpaper=true,colorlinks,bookmarks=false]{hyperref}

\wacvfinalcopy 

\def\wacvPaperID{0023} 

\ifwacvfinal\fi
\begin{document}

\title{Quantitative Analysis of Automatic Image Cropping Algorithms:\\A Dataset and Comparative Study}

\author{Yi-Ling Chen$^{1,2}$~~~~Tzu-Wei Huang$^3$~~~~Kai-Han Chang$^2$~~~~Yu-Chen Tsai$^2$\\Hwann-Tzong Chen$^3$~~~~Bing-Yu Chen$^2$\\
$^1$University of California, Davis~~~~$^2$National Taiwan University~~~~$^3$National Tsing Hua University
}

\maketitle
\ifwacvfinal\thispagestyle{empty}\fi

\begin{abstract}
Automatic photo cropping is an important tool for improving visual quality of digital photos without resorting to tedious manual selection.
Traditionally, photo cropping is accomplished by determining the best proposal window through visual quality assessment or saliency detection.
In essence, the performance of an image cropper highly depends on the ability to correctly rank a number of visually similar proposal windows.
Despite the ranking nature of automatic photo cropping, little attention has been paid to learning-to-rank algorithms in tackling such a problem.
In this work, we conduct an extensive study on traditional approaches as well as ranking-based croppers trained on various image features.
In addition, a new dataset consisting of high quality cropping and pairwise ranking annotations is presented to evaluate the performance of various baselines.
The experimental results on the new dataset provide useful insights into the design of better photo cropping algorithms.
\end{abstract}

\section{Introduction}
\label{sec:intro}

\begin{figure}[h]
\centering
\begin{tabular}{c@{\hspace{0.1cm}}c@{\hspace{0.1cm}}c}
\includegraphics[height=5.5cm]{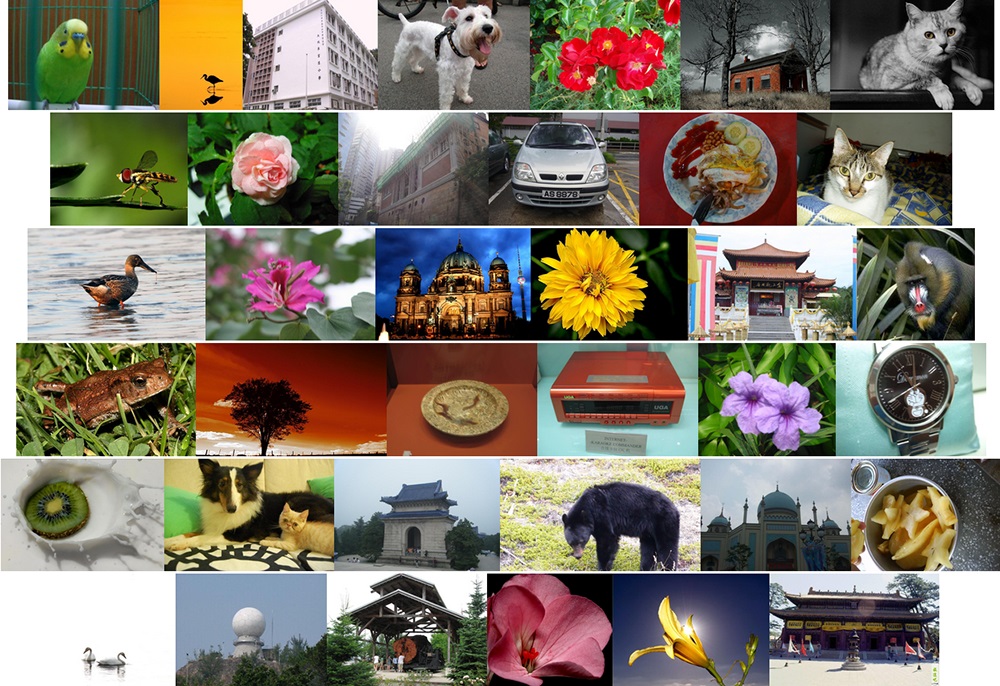}\\
\hline \\
\includegraphics[height=5.5cm]{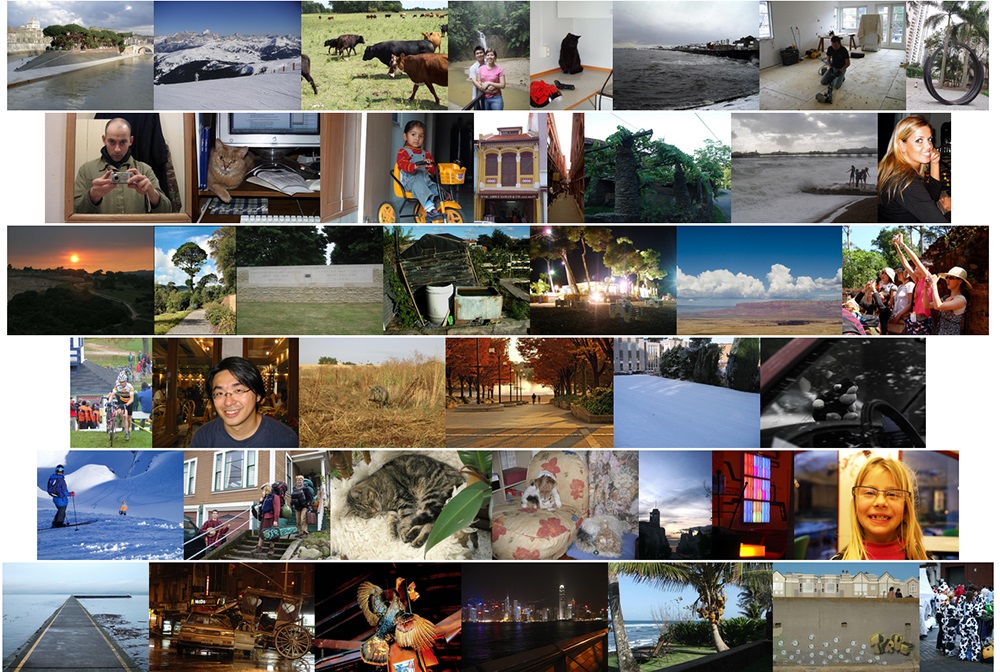}
\end{tabular}
\caption{A new image cropping dataset is presented in this work. Example images of an existing database \cite{Yan:CVPR:2013} (upper) and ours (bottom). Note that \cite{Yan:CVPR:2013} contains more images of canonical views. We attempt to build a new dataset containing more images of non-canonical perspectives and richer contextual information.}
\end{figure}

Photo cropping is an important operation for improving visual quality of photos, which is mainly performed to remove unwanted scene contents or irrelevant details by cutting away the outer parts of the image.
Nowadays, it is mostly performed on digital images but remains a tedious manual selection process and requires experience to obtain quality crops.
Therefore, a lot of computational techniques have been proposed to automate this process \cite{Zhang:2013:TIP,Yan:CVPR:2013,Fang:MM:2014,Zhang:2014:TIP:1}. 

Automatic photo cropping is closely related to other applications like image thumbnail generation \cite{Suh:UIST:2003,Marchesotti:ICCV:2009}, view finding and recommendation \cite{Chang:ICCV:2009,Cheng:MM:2010,Su:TMM:2012}.
In a nutshell, these approaches share one core capability in common -- finding an optimal subview in terms of its \emph{aesthetics} or \emph{composition} within a larger scene.
In other words, their performance highly depends on the ability to correctly \emph{rank} a number of visually similar proposal windows.
Traditionally, automatic photo cropping techniques follow two mainstreams, \ie \emph{attention}-based \cite{Stentiford:ICVS:2007} and \emph{aesthetics}-based methods \cite{Nishiyama:MM:2009}, which aim to search for a crop window covering the most visually significant objects or assess the visual quality of the candidate windows according to certain photographic guidelines, respectively.
However, in spite of its nature of a ranking problem, to the best of our knowledge none of the existing researches have adopted the \emph{learning-to-ranking} approaches to accomplish this task, which is proven to be useful and widely used in many information retrieval systems.
The main goal of this work is thus to study the effectiveness of applying ranking algorithms on image cropping and view finding problems.

We believe that the ability of ranking pairwise views \emph{in the same context} is essential for evaluating photo cropping techniques.
Therefore, we build a new dataset consisting of 1,743 images with human labeled crop windows and 31,430 pairs of subviews with visual preference annotations.
To obtain quality annotations, we carefully designed an image collection and annotation pipeline which extensively exploited a crowd-sourcing platform to validate the annotated images.
We conduct extensive evaluation on traditional approaches and a variety of machine learned rankers trained on the AVA dataset \cite{Murray:CVPR:2012} and our dataset with various aesthetic features \cite{Marchesotti:ICCV:2011,Donahue:2013:arXiv}.
Experimental validations show that ranking based image croppers consistenly achieve higher cropping accuracy in both the image cropping dataset \cite{Yan:CVPR:2013} and our dataset.
Additionally, it also suggests that ranking-based algorithms still have great potential to further improve their performance on automatic image cropping with more effective features.
The dataset presented in this work is publicly available\footnote{\url{https://github.com/yiling-chen/flickr-cropping-dataset}}.

\section{Previous Work}

\subsection{Aesthetic Assessment and Modeling}

The main goal of aesthetic visual analysis is to imitate human interpretation of the beauty of natural images.
Traditionally, aesthetics visual analysis mainly focuses on the binary classification problem of predicting high- and low-quality images \cite{Datta:ECCV:2006,Dhar:CVPR:2011,Luo:ICCV:2011}.
To this end, researchers design various features to capture the aesthetic properties of an image compliant with photographic rules or practices, such as the rule of thirds and visual balance.
For example, the spatial distribution of edges is exploited as a feature to model the photographic rule of ``\emph{simplicity}'' \cite{Ke:CVPR:2006}.
Some photo \emph{recomposition} techniques attempt to enhance image composition by rearranging the visual elements \cite{Bhattacharya:MM:2010}, applying crop-and-retarget operations \cite{Liu:EG:2010} or providing on-site aesthetic feedback \cite{Yao:2012:IJCV} to improve the aesthetics score of the manipulated image.

Instead of using ``hand-crafted'' features highly related to the best photographic practices, Marchesotti \etal show that generic image descriptors previously used for image classification are also capable of capturing aesthetic properties \cite{Marchesotti:ICCV:2011}.
In \cite{Isola:TPAMI:2014}, Isola \etal show that the \emph{memorability} of images is predictable by using global image descriptors.
In recent years, deep convolutional neural network (DCNN) has been proven to gain tremendous success in various visual recognition tasks and several works also exploited it as the machinery to learn effective features for aesthetics prediction \cite{Kang:CVPR:2014,Lu:MM:2014,Lu:ICCV:2015}.
In \cite{Karayev:BMVC:2014}, Karayev \etal compare the performance of different image features for style recognition and show that CNN features generally outperform other features even when trained on object class labels.

\subsection{Photo Cropping and View Finding Methods}

Generally, automatic photo cropping techniques can be categorized into two lines of researches: attention-based and aesthetics-based approaches. 
The basic principle of attention-based methods is to place the crop window over the most visually significant regions in an image according to certain attention scores, \eg saliency map \cite{Suh:UIST:2003,Stentiford:ICVS:2007}, or by resorting to eye tracker \cite{Santella:CHI:2006}, face detector \cite{Zhang:ICME:2005} to find the regions of interest.
In \cite{Marchesotti:ICCV:2009}, a classifier trained on an annotated database for saliency prediction is used to facilitate image thumbnail extraction.
Recently, Chen \etal \cite{Chen:CVPR:2016} conduct a complexity study of several different formulations of optimal window search under the attention-based framework. 
Although the attention-based approaches can usually determine a crop receiving the most human attention, the cropped images are not necessarily visually pleasing due to little consideration of image composition.

The aesthetics-based methods accomplish the cropping task mainly by analyzing the attractiveness of the cropped image with the help of a quality classifier \cite{Nishiyama:MM:2009,Fang:MM:2014}, and are thus closely related to photo quality assessment \cite{Datta:ECCV:2006,Dhar:CVPR:2011,Luo:ICCV:2011}. 
Recently, Yan \etal \cite{Yan:CVPR:2013} proposed a cropping technique that employs features designed to capture the changes between the original and cropped images.
In \cite{Chang:ICCV:2009}, finding good subviews in a panoramic scene is achieved by analyzing the structural features and layout of visual saliency learned from reference images of professional photographs.
Several other related works achieve view finding by learning aesthetic features based on position relationships between regions \cite{Cheng:MM:2010} or image decomposition \cite{Su:TMM:2012}.

\begin{figure*}[t]
\centering
\includegraphics[width=16cm]{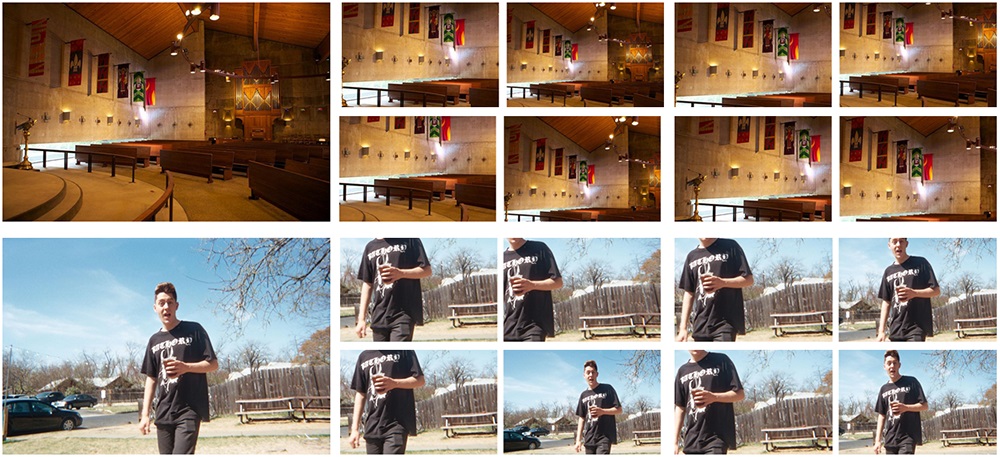}
\caption{Examples of crop pair generation. On the left are the source images and four corresponding crop pairs are shown on the right. Each pair of crop windows were randomly generated with the guidance of a saliency map to prevent from too much unimportant contents. The aesthetics preference relationship between the crop pairs were determined by the ranking results from AMT workers.}
\label{fig:ranking_pairs}
\end{figure*}

\subsection{Datasets for Computational Aesthetics}

Datasets play an important role for computer vision researches since they provide a means to train and evaluate algorithms. 
There are already several publicly available databases containing aesthetic annotations, such as \cite{Datta:ECCV:2006,Ke:CVPR:2006,Luo:ICCV:2011,Murray:CVPR:2012}.
Among them, AVA \cite{Murray:CVPR:2012} is a large-scale dataset which takes advantage of community-shared data (\eg \texttt{dpchallenge}) to provide a rich collection of aesthetic and semantic annotations.
Despite all these efforts, there is still not a standard benchmark for evaluating automatic photo cropping algorithms.
In \cite{Yan:CVPR:2013}, the authors built a dataset consisting of 950 images, which are divided into seven categories and individually cropped by three professional experts. 
We see two deficiencies of this dataset.
First, the selected image are from a database originally for photo quality assessment.
Some images of professional quality and compositions are also included and cropped.
These crops may not faithfully reflect the unwanted regions of the images.
Second, many images in the database are \emph{iconic} object or scene images which were taken from a canonical perspective, particularly in the \texttt{animal}, \texttt{architecture}, \texttt{static} categories, which may lack non-canonical views and contextual information.
To provide a more general benchmark, we choose to build a new dataset from scratch with a carefully designed image collection and annotation pipeline.

\section{Dataset Construction}

In this section we describe how the candidate images are selected and the design principles of the annotation pipeline.

\subsection{Design Principles}

While designing the image annotation procedure, a pilot study was carried out among the authors.
We randomly downloaded a small number of test images and had the authors to annotate the ideal crop windows individually.
Several observations were obtained after the pilot study.
\begin{enumerate}
\item Photo cropping is sometimes very subjective. Particularly, it is extremely difficult to define an appropriate crop for photos of both professional and poor quality since there are no ``obvious'' answers.
\item Most online images are post-processed which means that most unwanted regions had been already cut away before they were uploaded.
Therefore, it is essential to search for ``raw'' images for annotation.
\item Sometimes people do agree others' crops are good even though they are different from their own crops.
To obtain quality crops, we decided to resort to a crowd-sourcing platform to review all the cropping annotations and adopt only the highly ranked ones as final results.
\end{enumerate}

In manual image cropping, human typically iterates the procedure of moving and adjusting the position, size and aspect ratio of the crop window, and examining the visual quality before and after the manipulation until an ideal crop window is obtained. 
It is essentially a problem of ranking a number of pairwise subviews that are visually similar.
Inspired by the aforementioned process, we also build annotations indicating such preference relationships.
We believe that this type of data will be beneficial for researchers to more faithfully evaluating the performance of image cropping techniques.

\subsection{Image Collection}

In the image collection stage, we aimed to collect as many non-iconic images as possible for better generalization capability \cite{Torralba:CVPR:2011}.
Following the strategy suggested in \cite{Lin:ECCV:2014}, we chose Flickr as our data source, which tends to have fewer iconic images.
In addition, we searched Flickr with many combinations of a pre-defined set of keywords, by which more non-iconic images with richer contextual information are more likely to be returned.
The above process resulted in an initial set of candidate images consisting of 31,888 images, which were then passed through a data cleaning process.
We employed workers on Amazon Mechanical Turk (AMT) to filter out inappropriate images, such as collage, computer-generated images or images with post-processed frames.
Particularly, we also asked the AMT workers to pick the photos of excellent quality, because they are potentially not necessary for cropping and thus not suitable for annotation.
After data cleaning, 18,925 images remained to enter the next stage of data annotation.

\subsection{Image Annotation}

We collected two types of annotation through crowd-sourcing in our dataset.
\begin{itemize}
\item \textbf{Cropping annotation:} 
We built a web-based interface for performing the image cropping tasks.
The users were recruited from the photography club in our university by invitation.
In our task design, we allowed users to skip the images which were judged to be unnecessary for cropping.
We eventually retrieved 3,413 cropped images after the human labeling process was finished.
For validation, we grouped pairs of cropped image and its corresponding source image as Human Intelligence Tasks (HITs) and assigned each of them to 7 distinct workers on AMT.
It is worth noting that a qualification test consisting of 10 pictorial ranking questions was given to each worker.
Only the workers who correctly answered at least 8 questions were allowed to take the HITs.
For each HIT, the order that the source and cropped image appeared in the HIT was randomized and the workers were asked to pick the more preferable one in terms of their aesthetics.
In total, 1,743 out of the 3,413 cropped images were ranked as preferable by at least 4 workers and they constitute the final cropping annotation of our dataset.

\item \textbf{Ranking annotation:} 
Besides the cropping annotation, we want to enrich the dataset with pairwise ranking relationships between subviews in the same image.
For each image with human labeling, 10 pairs of crop windows were randomly generated and then ranked by 5 workers with a similar process as the cropping annotations.
To prevent the crop windows from containing too much unimportant contents, we utilized a saliency map \cite{Vig:CVPR:2014} to guide crop selection.
The size of crop windows varied to imitate the effect of zoom in/zoom out and each pair of crop windows possessed sufficient overlapping.
\figname\ref{fig:ranking_pairs} illustrates some examples of the generated crop pairs for ranking.
We eventually obtained a collection of totally 34,130 pairs of crop windows with aesthetics preference information.
Note that the human cropped images and the corresponding source images can also be treated as ranking annotations. 
\end{itemize}

To summarize, our dataset is composed of 3,413 cropped images and 34,130 crop pairs generated from the corresponding images.
All the source/crop and crop/crop pairs were reviewed by a number of human workers to derive the pairwise aesthetics relationships as the ranking annotation.
Finally, 1,743 out of the 3,413 human cropped images were selected as the final cropping annotation of our dataset.


\section{Algorithmic Analysis}

In this section, we first describe the experimental settings and baseline algorithms, and then demonstrate the experimental validation results.

\subsection{Experimental Settings}

\subsubsection{Datasets}

We adopt both the AVA \cite{Murray:CVPR:2012} and our dataset to train various image croppers to be compared in this study.
The average aesthetic score associated with each image in AVA are used to select a set of high and low quality images to train a photo quality classifier (Section \ref{sec:aesthetic_method}).
Additionally, we also exploit the aesthetic scores to form relative ranking constraints to train ranking-based image croppers (Section \ref{sec:ranking_methods}).

For our dataset, we split the cropping and ranking annotations into training and test set with a roughly 4:1 ratio.
Specifically, 348 out of the 1,743 images with highly ranked crops are adopted as ground truth for evaluating the performance of image croppers.
The ranking annotations are also used to train ranking-based image croppers (Section \ref{sec:ranking_methods}).

Finally, the image cropping annotations in \cite{Yan:CVPR:2013} is also used to evaluate the performance of image croppers.

\subsubsection{Evaluation Protocol}
\label{sec:protocol}

For fair comparison, we take the strategy of evaluating all baseline algorithms on a number of sliding windows. 
For simplicity, we set the size of search window to each scale among $\left[0.5, 0.6, \dots, 0.9\right]$ of the original image and slide the search window over a 5$\times$5 uniform grid.
The optimal crop windows determined by image croppers are compared to the ground truth to evaluate their performance.

We adopt the same evaluation metrics as in \cite{Yan:CVPR:2013}, \ie, \emph{average overlapped ratio} and \emph{average boundary displacement error} to measure the \emph{cropping accuracy} of image croppers.
The average overlapped ratio is computed by
\begin{equation}
\frac{1}{N}\sum_{i=1}^{N} area(W^g_i \cap W^c_i)/area(W^g_i \cup W^c_i),
\end{equation}
where $W^g_i$ and $W^c_i$ denote the ground-truth crop window and the crop window determined by the baseline algorithms for the $i$-th test image, respectively. 
$N$ is the number of test images.
The boundary displacement error is given by
\[
\sum_{j=\{l,r,b,u\}} ||B^g_j-B^c_j||/4,
\]
where $B^g_i$ and $B^c_i$ denote the four corresponding edges between $W_g$ and $W_c$.
Note that the boundary displacements have to be normalized by the width or height of the original image. 

We optionally report the \emph{swap error} evaluated on the test set of AVA and our dataset. 
It is the ratio of swapped pairs averaged over all queries, which measures the \emph{ranking accuracy} of image croppers to correctly rank pairwise subviews.

\subsubsection{Aesthetic Features}
\label{sec:aesthetic_feature}

For all the learning-based image croppers, we adopt the ``deep'' activation features \cite{Donahue:2013:arXiv} to accomplish aesthetics prediction as suggested in \cite{Karayev:BMVC:2014}.
For feature extraction, we exploit the implementation of AlexNet \cite{Krizhevsky:NIPS:2012} provided by the Caffe library \cite{jia2014caffe}.
Each training sample is resized to 227-by-227 pixels and forward-propagated into the network.
The activations of the last fully-connected layer are retained as the aesthetic features ($\mathbf{DeCAF}_7$), which are of 4,096-dimension.

We optionally train the ranking-based image croppers with generic image descriptors \cite{Marchesotti:ICCV:2011} to inspect the performance variations.
Specifically, Fisher vectors of SIFT descriptors with spatial pyramid (\textbf{SIFT-FV}) and Fisher vectors of color descriptors with spatial pyramid (\textbf{Color-FV}) are considered.
For \textbf{SIFT-FV} and \textbf{Color-FV}, the cardinality of visual words is 256, and the image descriptor is constructed by concatenating the features extracted from 8 sub-image layouts: ``1$\times$1'' (whole image), ``3$\times$1'' (upper, center, bottom), ``2$\times$2'' (quadrant). 
The feature points are densely evaluated every 4 pixels, resulting in 262,144-dimension feature vectors.

\subsection{Baseline Algorithms}

\subsubsection{Attention-Based Methods}

The first category of methods to be compared are the extension of the \emph{attention-based} photo cropping methods \cite{Suh:UIST:2003,Stentiford:ICVS:2007}, which take advantage of the saliency map accompanying the original image to search for an optimal crop window with the highest average saliency. 
Instead of the outdated saliency detection methods used in the previous works, we adopt two state-of-the-art methods, \ie, \textbf{BMS} \cite{Zhang:ICCV:2013} and \textbf{eDN} \cite{Vig:CVPR:2014}, with leading performance on the CAT2000 dataset from MIT Saliency Benchmark \cite{mit-saliency-benchmark}. 
In addition to the aforementioned search strategy (\textbf{MaxAvg}), we further implement another search criterion, which maximizes the difference of average saliency between the crop window and the outer region of the image (\textbf{MaxDiff}).
The saliency maps are generated by the implementation of the original authors with the default parameter settings.

\subsubsection{Aesthetics-Based Method}
\label{sec:aesthetic_method}

The second category of comparison techniques represent the research line of \emph{aesthetics-based} methods, which exploit a quality classifier that measures whether the cropped region is visually attractive to users \cite{Nishiyama:MM:2009,Fang:MM:2014}.
Instead of the low-level features used in the previous works, we adopt the more advanced $\mathbf{DeCAF}_7$ features \cite{Donahue:2013:arXiv} to achieve aesthetics recognition.
A total of 52,000 images with the highest and lowest aesthetics scores are selected from the AVA dataset \cite{Murray:CVPR:2012} as the training (67\%) and testing (33\%) samples.
We thus train a binary SVM classifier with RBF kernels, which predicts a photo as high or low quality.
The parameters of the classifier are obtained through 5-fold cross validation on the training set and the testing accuracy achieved 80.27\%.
To use the binary classifier as an image cropper, we take advantage of the method described in \cite{Lin:ML:2007} to compute the posterior class probability as the aesthetics score to pick the best crop among all candidate windows.

\subsubsection{Ranking-Based Methods}
\label{sec:ranking_methods}

\begin{table}
\begin{center}
\begin{tabular}{|c||c||c|}
\hline
Method                               & Overlap &  Disp.     \\ 
\hline\hline
RankSVM \cite{Joachims:KDD:2006}     & \textbf{0.6019}  &  0.1060    \\ 
\hline
RankNet \cite{Burges:ICML:2005}      & 0.6015  &  \textbf{0.1058}    \\ 
\hline
RankBoost \cite{Freund:2003}         & 0.5017  &  0.1383    \\ 
\hline
LambdaMART \cite{Wu:2010}            & 0.5451  &  0.1217    \\
\hline
\end{tabular}
\end{center}
\caption{Benchmarking of various learning-to-rank algorithms.
$\mathbf{DeCAF}_7$ feature is used to train the image rankers.
The cropping accuracy is evaluated on the 348 test images of our dataset.
The best results are highlighted in bold.}
\label{tab:ranking_algirithms}
\end{table}

The third category of comparison techniques are a family of aesthetics-aware image rankers. 
To choose an appropriate training algorithm, we have test several pairwise learning-to-ranking algorithms to train the image rankers, including RankSVM \cite{Joachims:KDD:2006}, RankNet \cite{Burges:ICML:2005}, RankBoost \cite{Freund:2003} and LambdaMART \cite{Wu:2010}.
We exploit the implementation of the above algorithms provided by SVM$^{rank}$\footnote{\url{https://www.cs.cornell.edu/people/tj/svm_light/svm_rank.html}} and RankLib\footnote{\url{https://people.cs.umass.edu/~vdang/ranklib.html}} libraries for our experiments.
The image rankers are trained by using the training set of our dataset with many different configurations of the individual algorithms.
The best-performing models are determined by 5-fold cross validation.
As summarized in \tabname~\ref{tab:ranking_algirithms}, RankSVM and RankNet achieve very competitive performance in terms of cropping accuracy.
However, since RankNet rankers take much longer time to train, we thus choose RankSVM as the training method for the rest of the experiments in this study.
Specifically, all the SVM rankers are trained with a linear kernel and use L1-norm penalty for the slack variables. 
The loss is measured by the total number of swapped pairs summed over all queries. 
The parameter $C$, which controls the trade-off between training error and margin, is determined via 5-fold cross validation.

\subsection{Evaluations and Analysis}

\emph{1) Comparison of traditional methods}:
As shown in \tabname~\ref{tab:cropping_performance}, the first five rows summarize the performances of the four variants of attention-based methods and the aesthetics-based method.
One can see that the search strategy of \textbf{MaxDiff} consistently outperforms \textbf{MaxAvg} for either type of saliency maps.
The possible reason is that \textbf{MaxDiff} tends to include more salient regions into the crop window in order to lower the total saliency score of the outer region.
Unlike \textbf{MaxAvg} which usually only concentrates on a single salient region, \textbf{MaxDiff} is more likely to obtain a crop window that forms a good composition. 

The performance of attention-based methods are highly dependent on the underlying saliency detection scheme.
Although \textbf{eDN} \cite{Vig:CVPR:2014} and \textbf{BMS} \cite{Zhang:ICCV:2013} possess comparable performance in \cite{mit-saliency-benchmark}, their performance greatly varied in image cropping.
It suggests that a standard benchmark is essential to choose the best saliency detection method for automatic image cropping.
A hybrid method that optimizes the compositional layout of salient objects might be less sensitive to the selection of saliency maps, such as \cite{Zhang:ICME:2005}.
In general, attention-based methods performed poorly in determining the aesthetics preferences between crop pairs (45.34\% -- 63.66\% swap error).
We believe that this phenomenon could be accounted for the lack of aesthetics considerations in this family of methods.
Note that the swap errors are calculated by the attention scores received by the crop pairs.

Comparing with attention-based methods, the aesthetics-based method (\textbf{SVM}+$\mathbf{DeCAF}_7$) achieved better performance in all evaluation metrics.
However, although the SVM classifier showed good capability of predicting high- and low-quality images, it did not perform well in ranking pairwise views (\ie, 42\% swap error), resulting in moderate performance in image cropping accuracy.

\emph{2) Comparison of various aesthetic features}:
The 9-th to 11-th rows of \tabname~\ref{tab:cropping_performance} compare the performance of image rankers trained by different aesthetic features using our new dataset.
$\mathbf{DeCAF}_7$ achieves the best accuracy in all metrics.
This result is consistent with the findings reported by \cite{Karayev:BMVC:2014}, \ie,
$\mathbf{DeCAF}_7$ generalizes well to other visual recognition tasks even though the DCNN was trained for object classification.
Although $\mathbf{SIFT}$-$\mathbf{FV}$ achieves comparable cropping accuracy with $\mathbf{DeCAF}_7$, the later obviously provides a much more compact feature representation of visual aesthetics.

\begin{table}[t]
\begin{center}
\begin{tabular}{|c||c||c||c|}
\hline
Method                                        & Overlap &  Disp.    &  Swap     \\ 
\hline\hline
\textbf{eDN} (\textbf{MaxAvg})                & 0.3573  &  0.1729   &  0.6366   \\ 
\hline
\textbf{eDN} (\textbf{MaxDiff})               & 0.4857  &  0.1372   &  0.4534   \\ 
\hline
\textbf{BMS} (\textbf{MaxAvg})                & 0.3427  &  0.1815   &  0.5775   \\ 
\hline
\textbf{BMS} (\textbf{MaxDiff})               & 0.3905  &  0.1674   &  0.4962   \\ 
\hline\hline
\textbf{SVM}+$\mathbf{DeCAF}_7$               & 0.5154  &  0.1325   &  0.4201   \\
\hline\hline
\textbf{AVA 1-1}+$\mathbf{DeCAF}_7$           & 0.5223  &  0.1294   &  \textbf{0.1317}   \\ 
\hline
\textbf{AVA 2-2}+$\mathbf{DeCAF}_7$           & 0.5069  &  0.1346   &  0.2379   \\ 
\hline
\textbf{AVA 5-5}+$\mathbf{DeCAF}_7$           & 0.4931  &  0.1384   &  0.2775   \\ 
\hline\hline
\textbf{Our}+$\mathbf{SIFT}$-$\mathbf{FV}$    & 0.5917  &  0.1084   &  0.3068   \\ 
\hline
\textbf{Our}+$\mathbf{Color}$-$\mathbf{FV}$   & 0.5042  &  0.1405   &  0.3692   \\
\hline
\textbf{Our}+$\mathbf{DeCAF}_7$               & \textbf{0.6019}  &  \textbf{0.1060}   &  0.3225   \\ 
\hline
\end{tabular}
\end{center}
\caption{Summarization of performance evaluation. 
The middle two columns measure the cropping accuracy on the 348 testing images of our dataset.
The best results are highlighted in bold.
}
\label{tab:cropping_performance}
\end{table}

\emph{3) Comparison of different datasets}:
In this experiment we examine the effectiveness of training image rankers on the AVA dataset \cite{Murray:CVPR:2012}.
Same as the aesthetics-based method, 52,000 images with the highest and lowest aesthetics scores are first selected.
A configuration of \textbf{AVA n-n} means that we repeatedly select $n$ images from the high- and low-ranked group, respectively, and generate all combinations of the selected images to form the ranking constraints.
Note that the characteristics of pairwise ranking constraints formed by AVA and our dataset are very different since AVA differentiates the visual preferences \emph{between} distinct images while our dataset ranks visually similar subviews \emph{within} the same images.

Row 6-8 in \tabname~\ref{tab:cropping_performance} give the performances of three rankers trained on AVA using $\mathbf{DeCAF}_7$ feature.
\textbf{AVA 1-1} performs best both in cropping and ranking accuracy.
However, surprisingly, increasing ranking constraints (\textbf{AVA 2-2} and \textbf{AVA 5-5}) caused the performance to considerably drop instead.
It indicates that only a sparse set of pairwise ranking constraints defined by the aesthetic scores are useful for image ranking and naively pairing images would not improve the ranking accuracy.
Besides, although \textbf{AVA n-n}+$\mathbf{DeCAF}_7$ rankers generally outperform the traditional methods in ranking accuracy, it does not reflect on their cropping capability.
For example, the cropping accuracy of \textbf{AVA 1-1}+$\mathbf{DeCAF}_7$ outperforms the best-performing traditional method (\ie, \textbf{SVM}+$\mathbf{DeCAF}_7$) with only an insignificant margin even though it has a much greater ranking accuracy.
One possible reason is that the training data of AVA do not reflect the visual preference among visually similar views, which is essential for image cropping.

Such observation can be further validated by comparing to the rankers trained on our dataset.
\textbf{Our}+$\mathbf{DeCAF}_7$ achieves significant improvement in cropping accuracy using the same feature.
It is also interesting to note that \textbf{Our}+$\mathbf{DeCAF}_7$ does not perform well in its ranking accuracy.
The reason for the low ranking accuracy could be explained as follows:
Since $\mathbf{DeCAF}_7$ is trained for the purpose of object recognition, it is thus very likely that the $\mathbf{DeCAF}_7$ features extracted from similar views containing the same ``object'' to be also similar.
The same phenomenon can also be observed in other aesthetic features, \ie, $\mathbf{SIFT}$-$\mathbf{FV}$ and $\mathbf{Color}$-$\mathbf{FV}$.
It suggests that there is still great potential to improve ranking-based image croppers by jointly learning the feature representation and semantically meaningful embedding of image similarity with DCNN \cite{Wang:CVPR:2014}  instead of directly using $\mathbf{DeCAF}_7$.

\begin{table}[t]
\begin{center}
(a)\\
\begin{tabular}{|c||c||c|}
\hline
Method                                        & Overlap &  Disp.    \\ 
\hline\hline
\textbf{eDN} (\textbf{MaxDiff})               & 0.4636  &  0.1578   \\ 
\hline
\textbf{SVM}+$\mathbf{DeCAF}_7$               & 0.5005  & 0.1444    \\
\hline
\textbf{AVA 1-1}+$\mathbf{DeCAF}_7$           & 0.5142  &  0.1399   \\ 
\hline
\textbf{Our}+$\mathbf{DeCAF}_7$               & \textbf{0.6643}  &  \textbf{0.092}  \\ 
\hline
\end{tabular}\\
(b)\\
\begin{tabular}{|c||c||c|}
\hline
Method                                        & Overlap &  Disp.    \\ 
\hline\hline
\textbf{eDN} (\textbf{MaxDiff})               & 0.4399  & 0.1651    \\ 
\hline
\textbf{SVM}+$\mathbf{DeCAF}_7$               & 0.4918  & 0.1483    \\
\hline
\textbf{AVA 1-1}+$\mathbf{DeCAF}_7$           & 0.5034  & 0.1443    \\ 
\hline
\textbf{Our}+$\mathbf{DeCAF}_7$               & \textbf{0.6556}  &  \textbf{0.095}  \\
\hline
\end{tabular}\\
(c)\\
\begin{tabular}{|c||c||c|}
\hline
Method                                        & Overlap &  Disp.    \\ 
\hline\hline
\textbf{eDN} (\textbf{MaxDiff})               & 0.437   & 0.1659    \\ 
\hline
\textbf{SVM}+$\mathbf{DeCAF}_7$               & 0.4882  & 0.1491    \\
\hline
\textbf{AVA 1-1}+$\mathbf{DeCAF}_7$           & 0.4939  & 0.147     \\ 
\hline
\textbf{Our}+$\mathbf{DeCAF}_7$               & \textbf{0.6439}  &  \textbf{0.099}  \\
\hline
\end{tabular}
\end{center}
\caption{Cross dataset validation.
(a)-(c) summarize the cropping accuracy of the best performing image croppers of each category shown in \tabname~\ref{tab:cropping_performance}, which are evaluated on the three different sets of annotations in the database of \cite{Yan:CVPR:2013}.
The best results are highlighted in bold. 
}
\label{tab:cross_dataset}
\end{table}

\emph{4) Cross-dataset validation}:
In this experiment, we select the best performing image croppers from each category shown in \tabname~\ref{tab:cropping_performance} and directly apply them on the image cropping databset of \cite{Yan:CVPR:2013}.
This dataset is composed of 950 images, which were annotated by three different users.
Since this dataset contains only cropping annotations, it is thus only used to evaluate the cropping accuracy.
A similar sliding window approach as described in Section \ref{sec:protocol} is adopted for evaluation.
As shown in \tabname~\ref{tab:cross_dataset}, \textbf{Our}+$\mathbf{DeCAF}_7$ consistently achieves the highest accuracy in all annotation sets, which further validates the effectiveness of ranking pairwise subviews in image cropping.
Note that higher cropping accuracy is reported in \cite{Yan:CVPR:2013}.
Since this is a comparative study, no optimization on the parameters of crop windows (\ie, $x$, $y$, $w$ and $h$) was performed for fair comparison.
We believe that the performance of the ranking based image croppers can be further enhanced by incorporating appropriate crop selection procedures.

\begin{figure*}
\centering
\begin{tabular}{c@{\hspace{0.1cm}}c@{\hspace{0.1cm}}c@{\hspace{0.1cm}}c@{\hspace{0.1cm}}c}
\includegraphics[width=3.5cm]{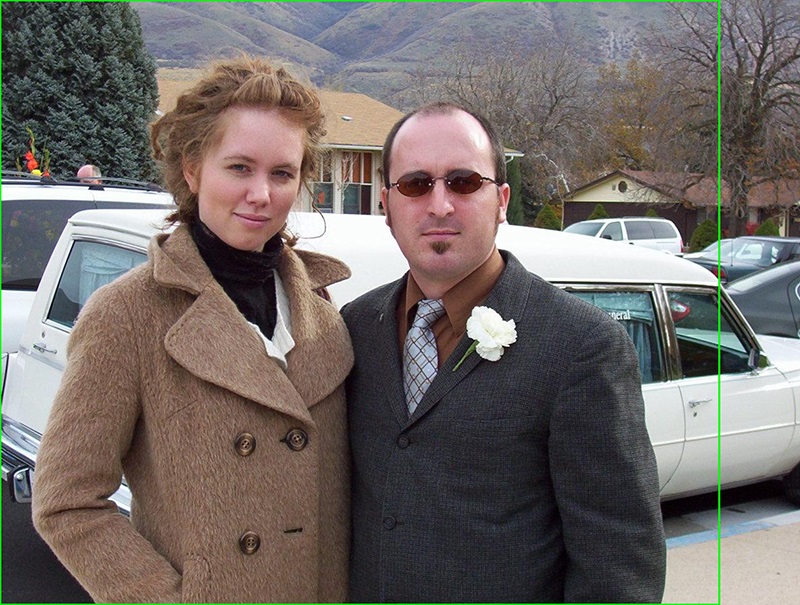}&
\includegraphics[width=3.5cm]{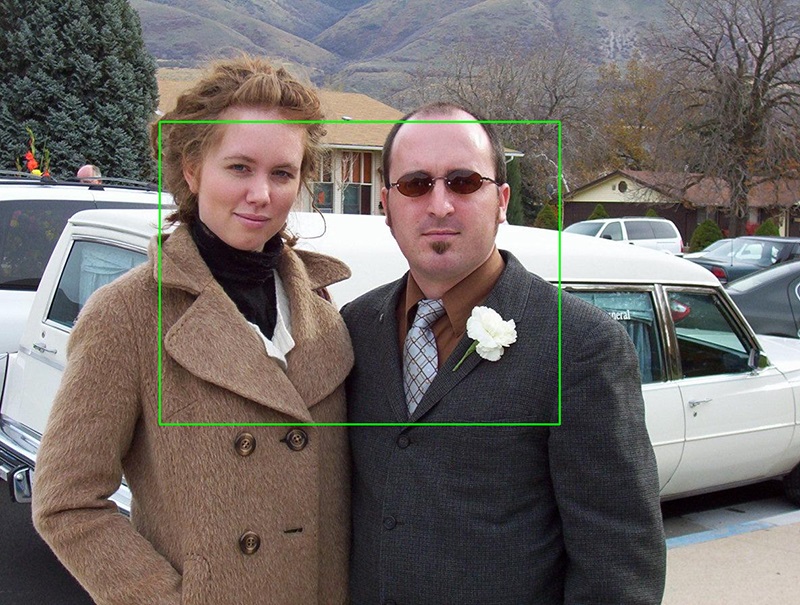}&
\includegraphics[width=3.5cm]{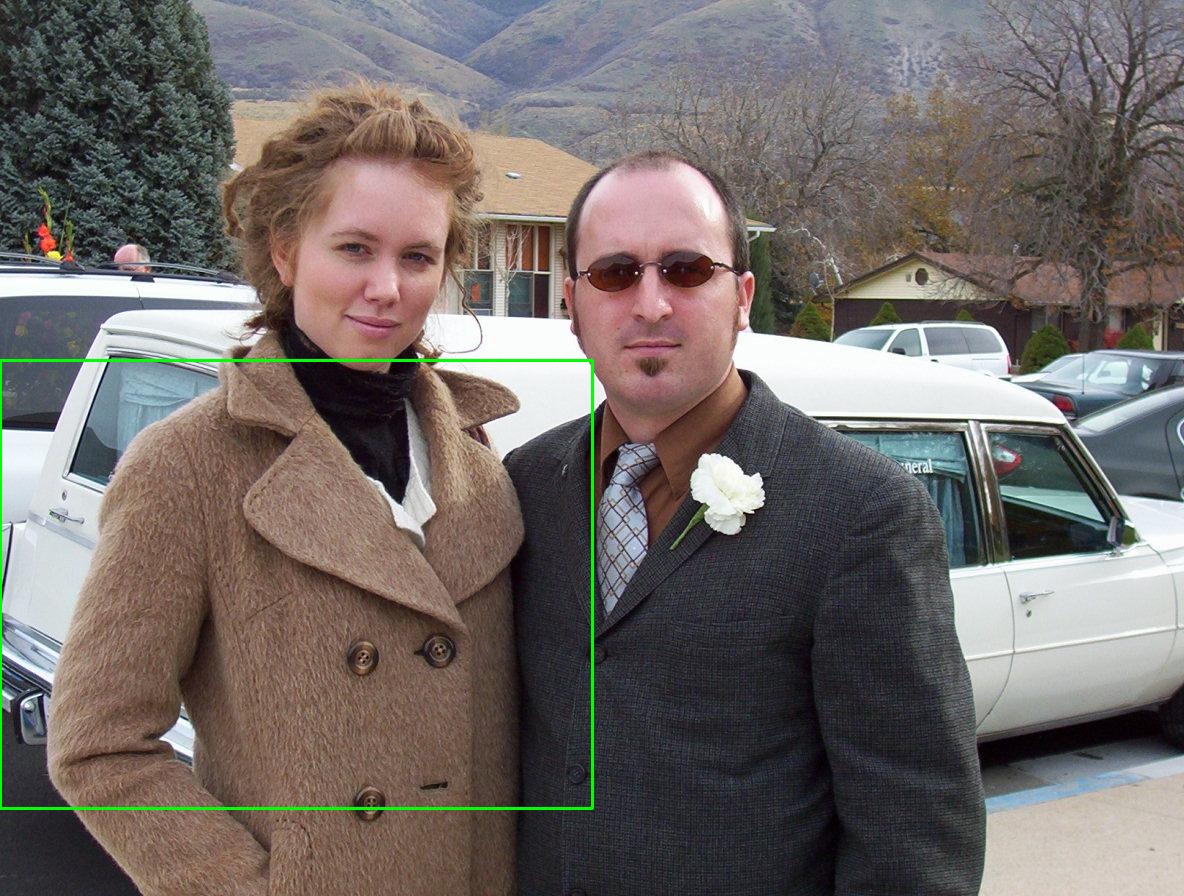}&
\includegraphics[width=3.5cm]{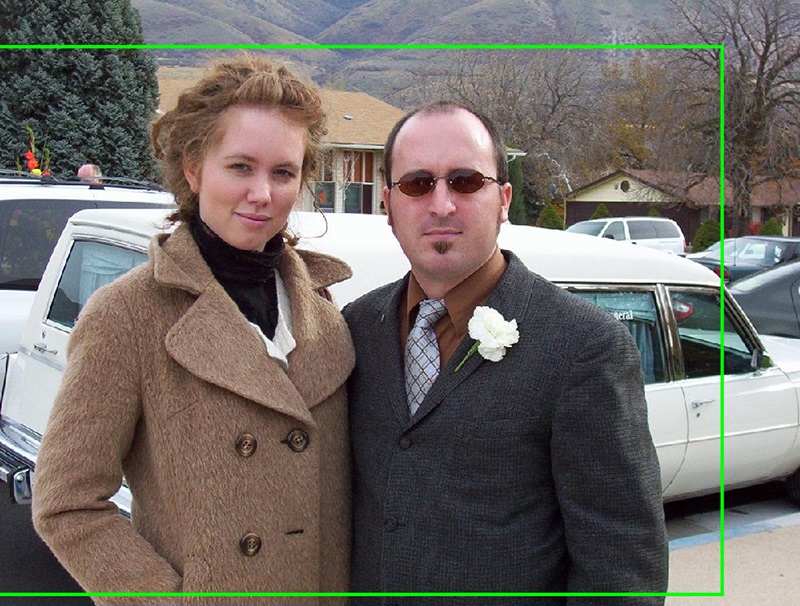}&\\

\includegraphics[width=3.5cm]{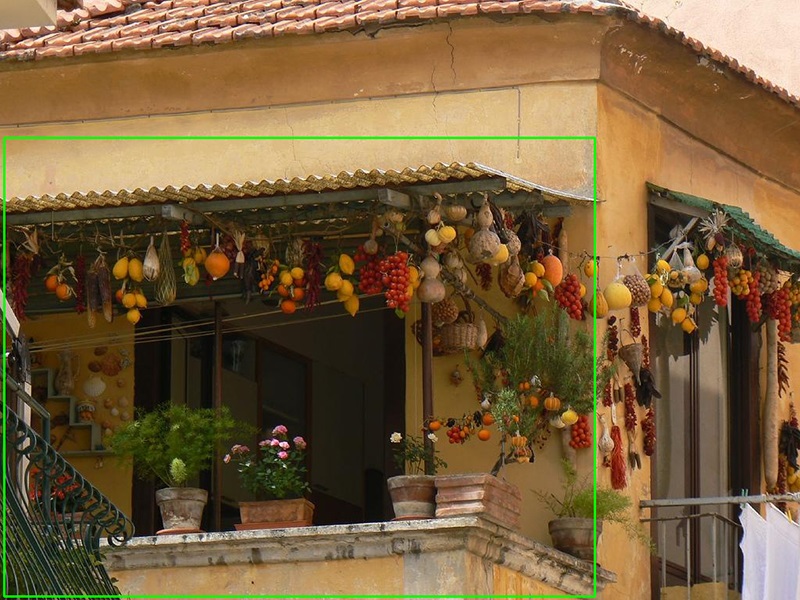}&
\includegraphics[width=3.5cm]{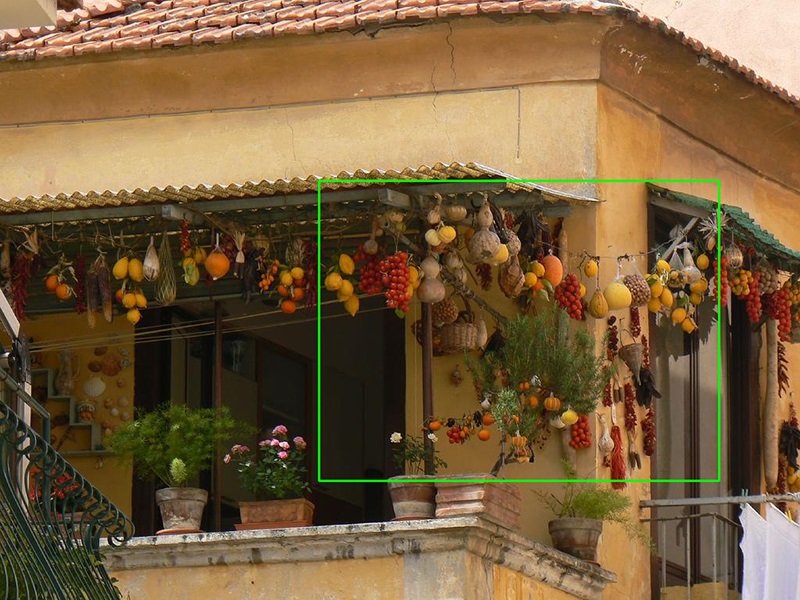}&
\includegraphics[width=3.5cm]{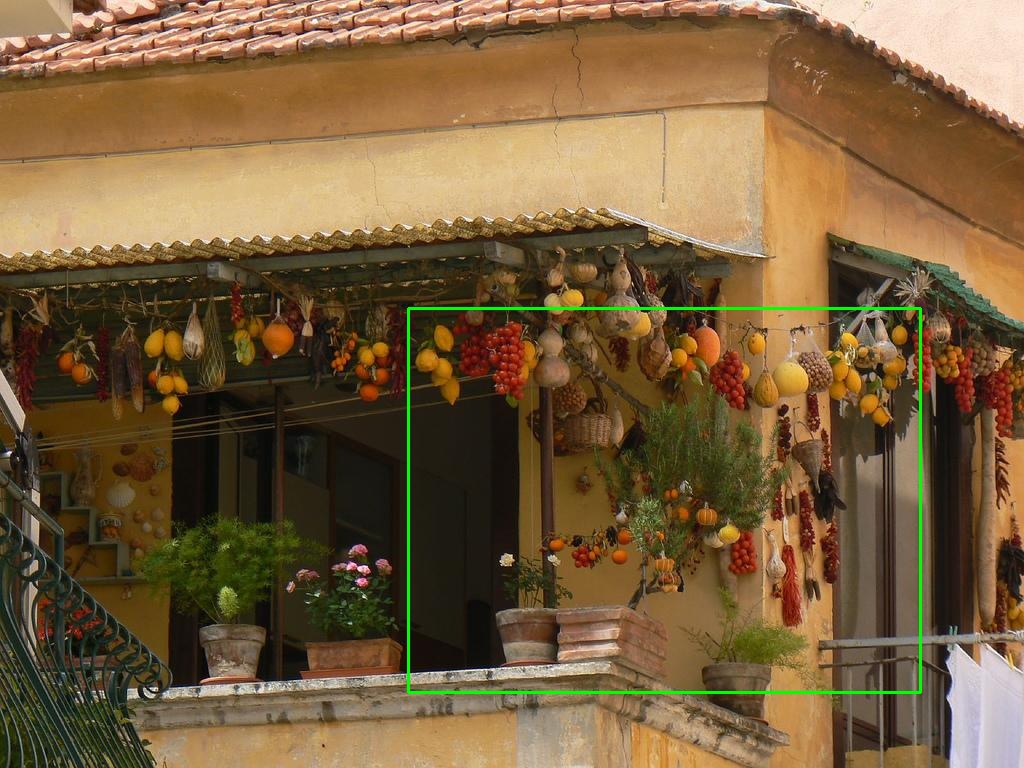}&
\includegraphics[width=3.5cm]{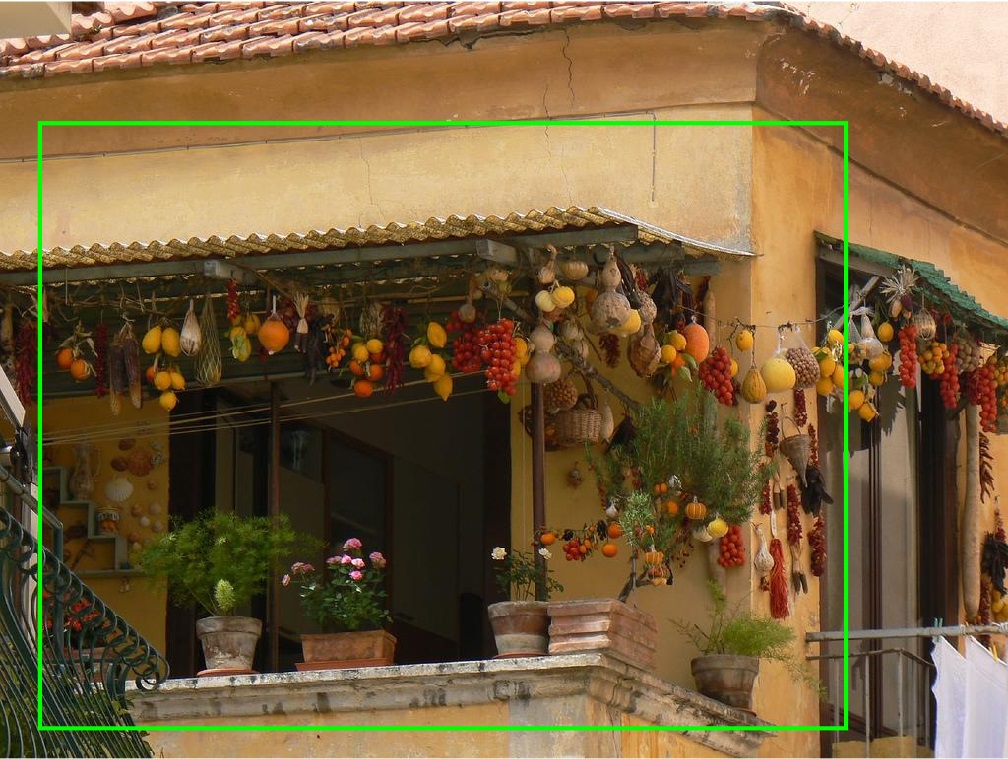}&\\

\includegraphics[width=3.5cm]{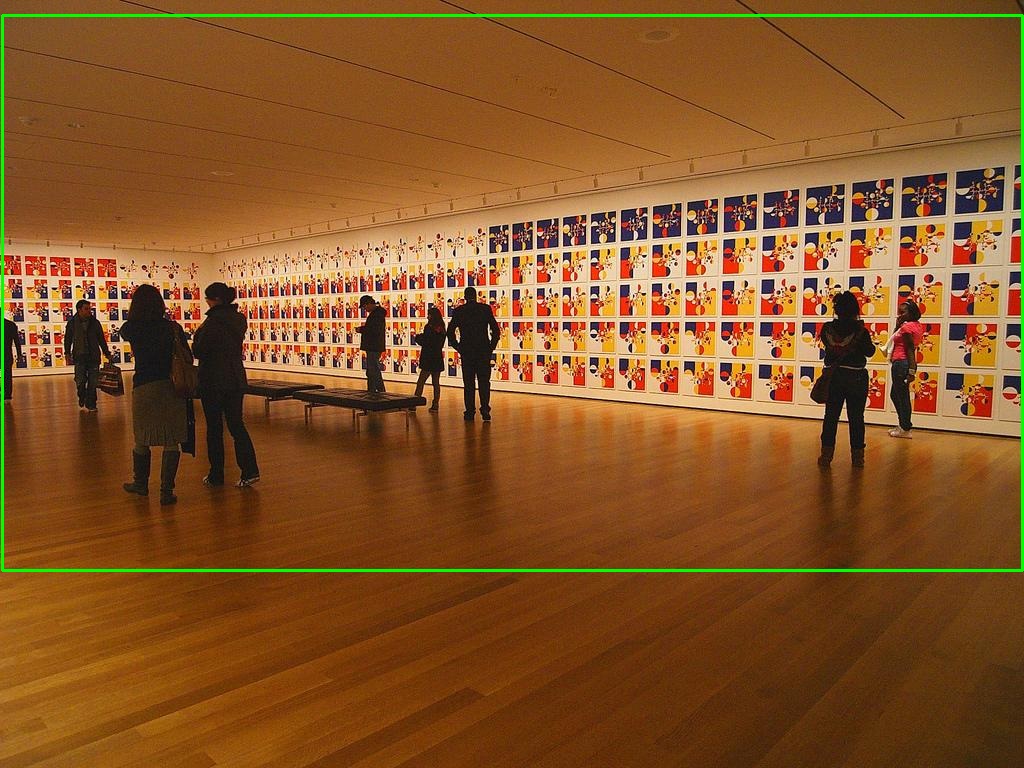}&
\includegraphics[width=3.5cm]{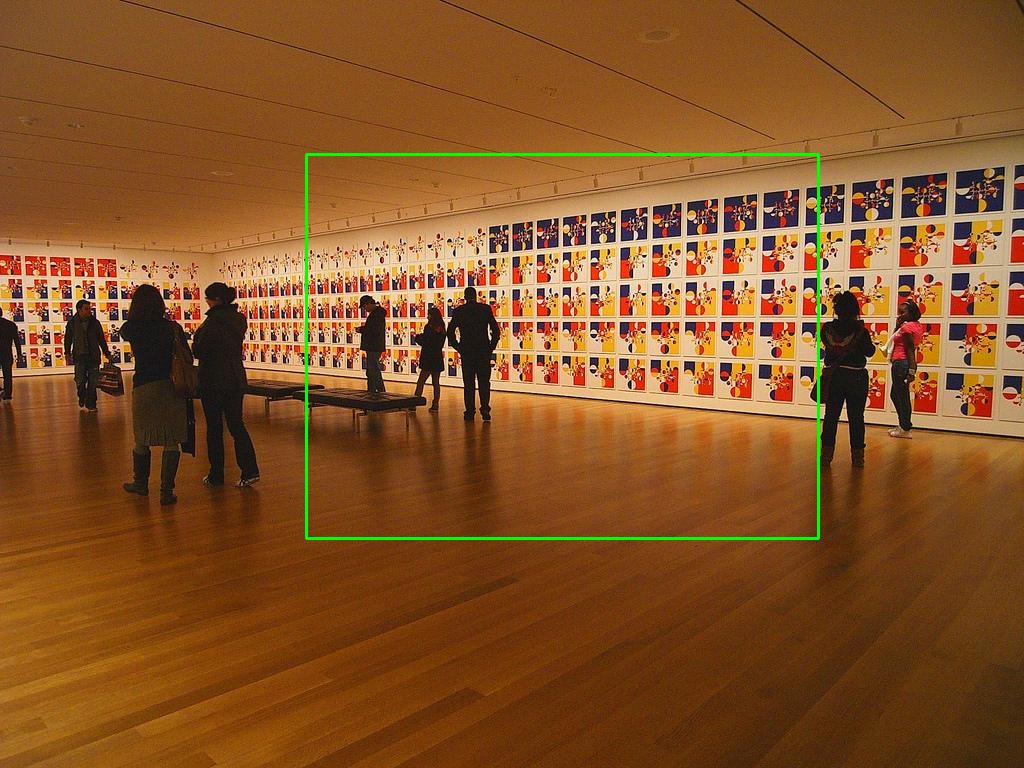}&
\includegraphics[width=3.5cm]{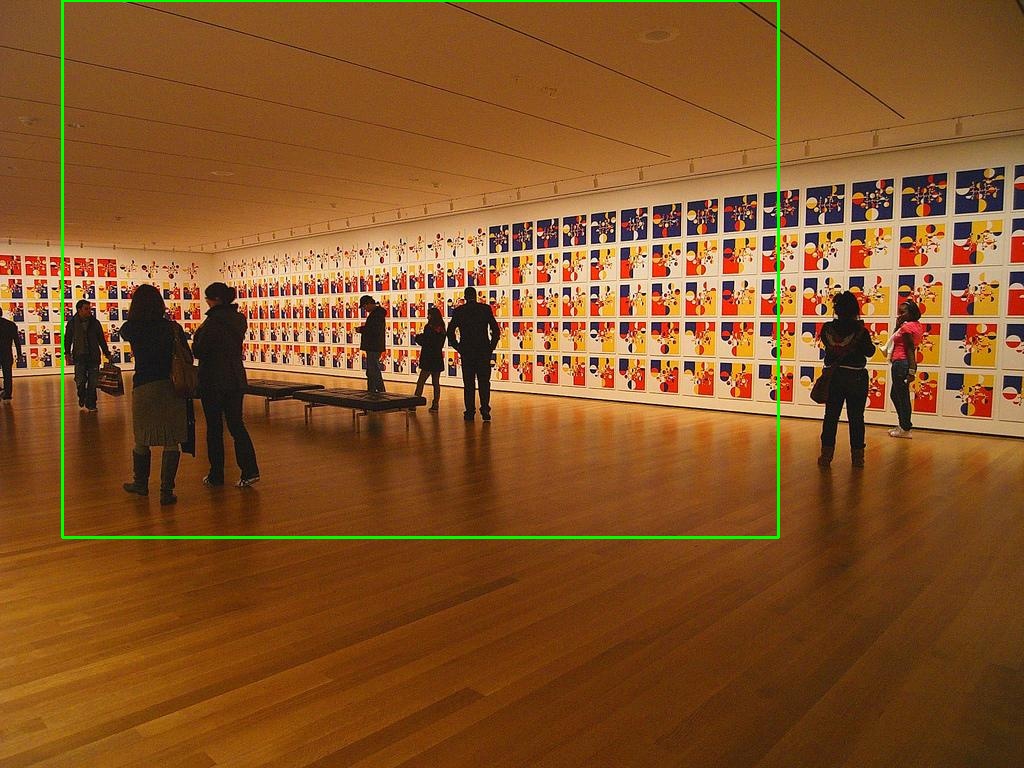}&
\includegraphics[width=3.5cm]{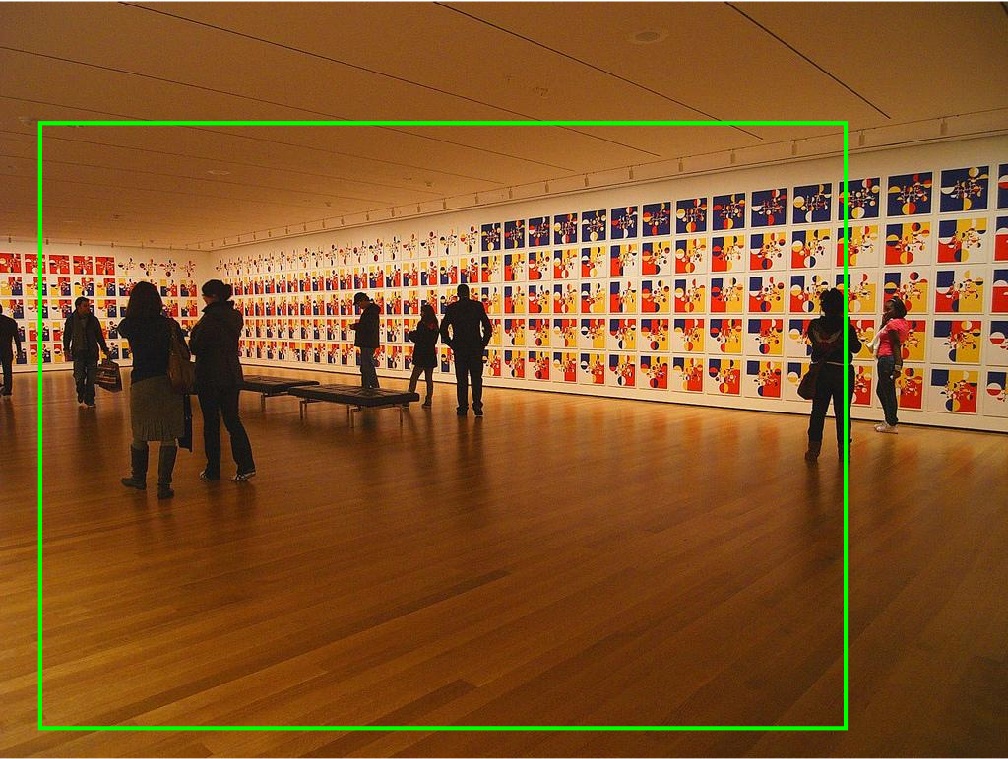}&\\

\includegraphics[width=3.5cm]{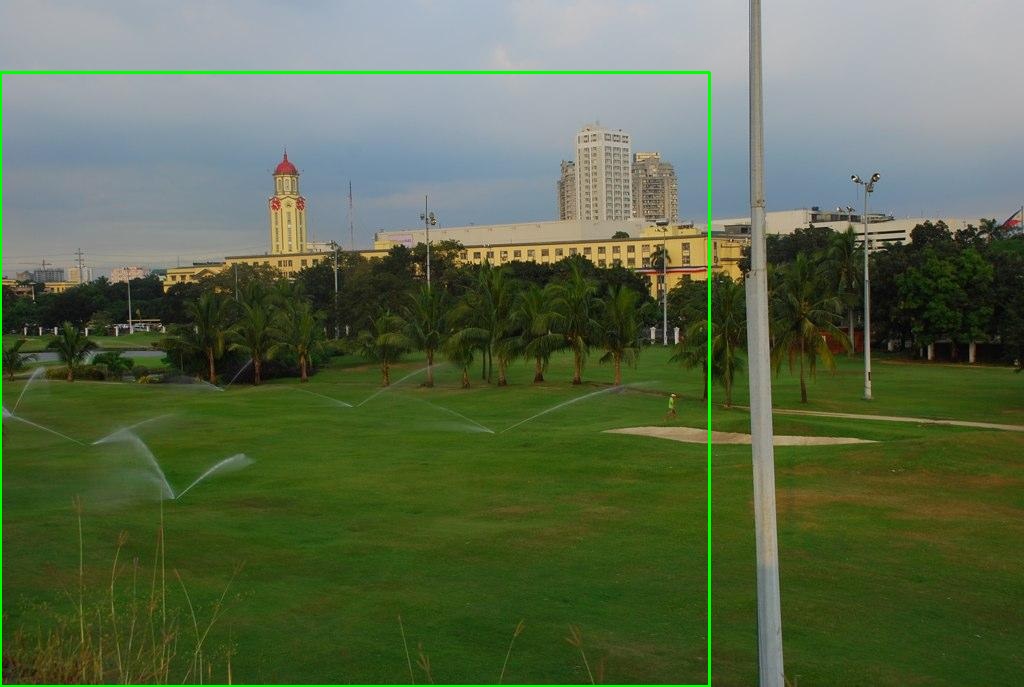}&
\includegraphics[width=3.5cm]{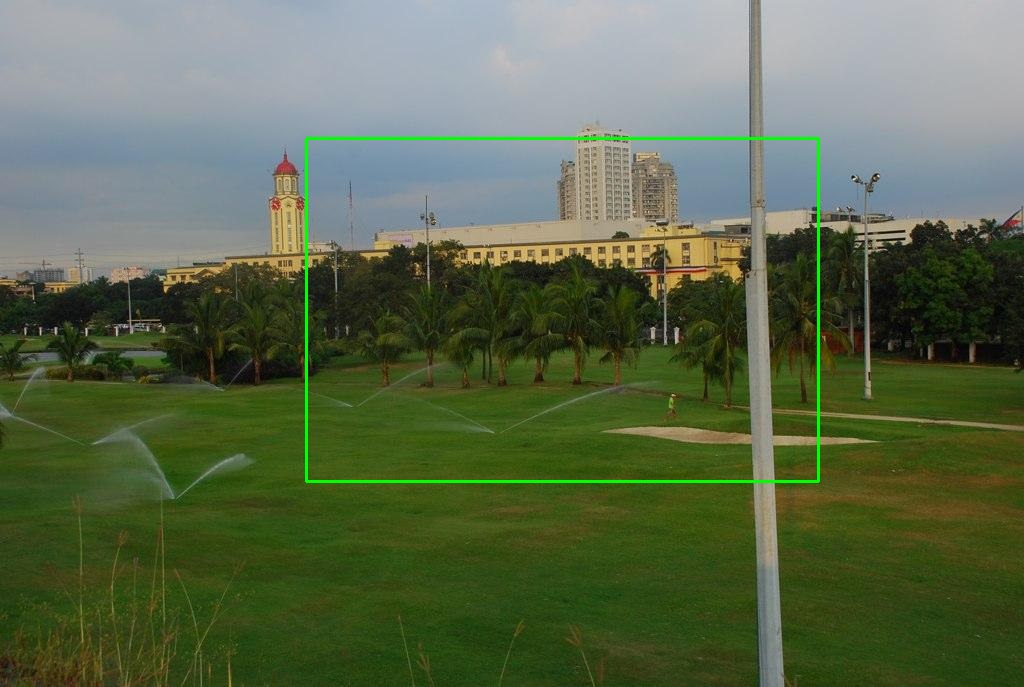}&
\includegraphics[width=3.5cm]{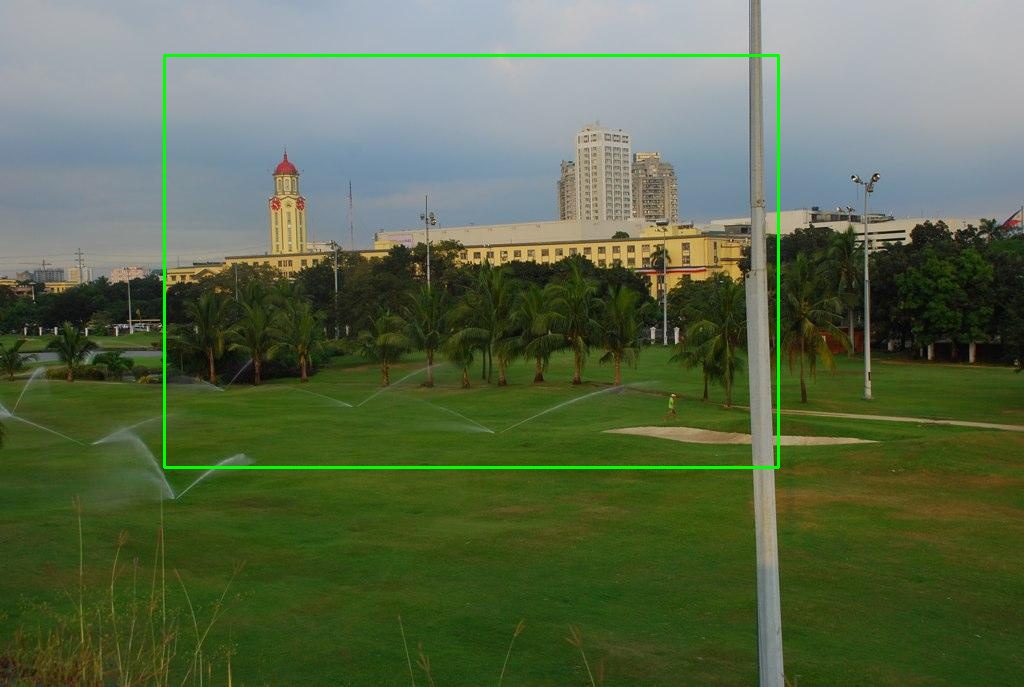}&
\includegraphics[width=3.5cm]{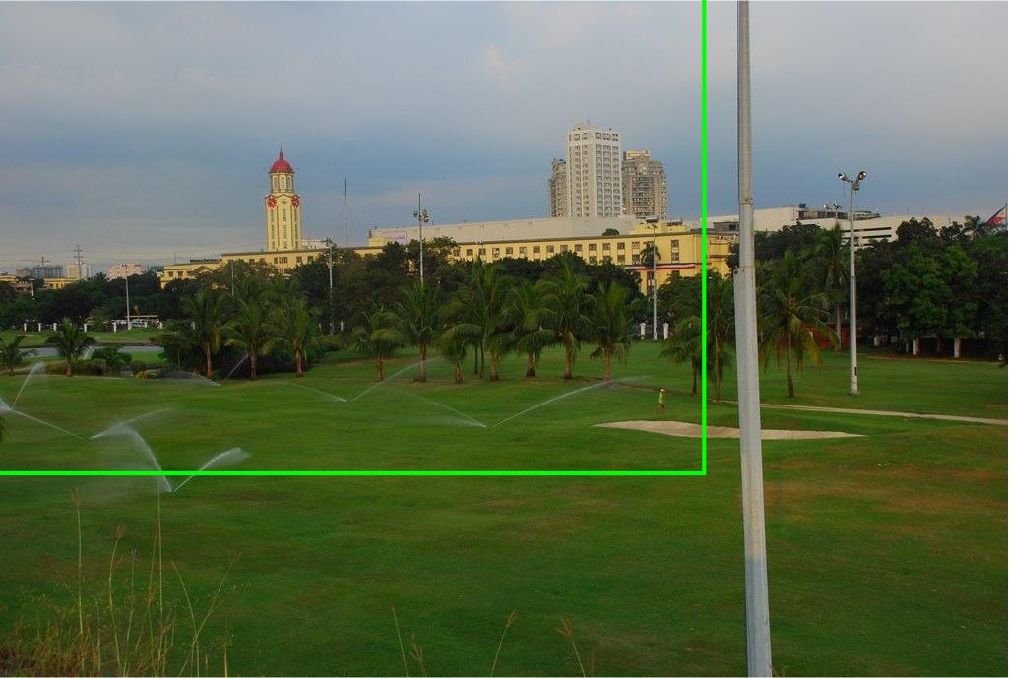}&\\

\includegraphics[width=3.5cm]{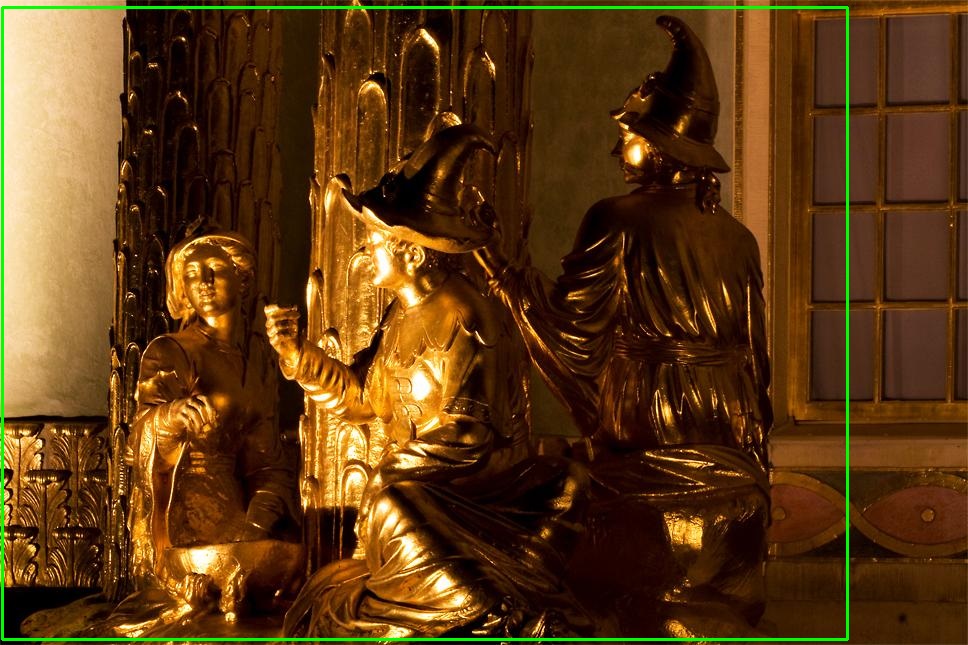}&
\includegraphics[width=3.5cm]{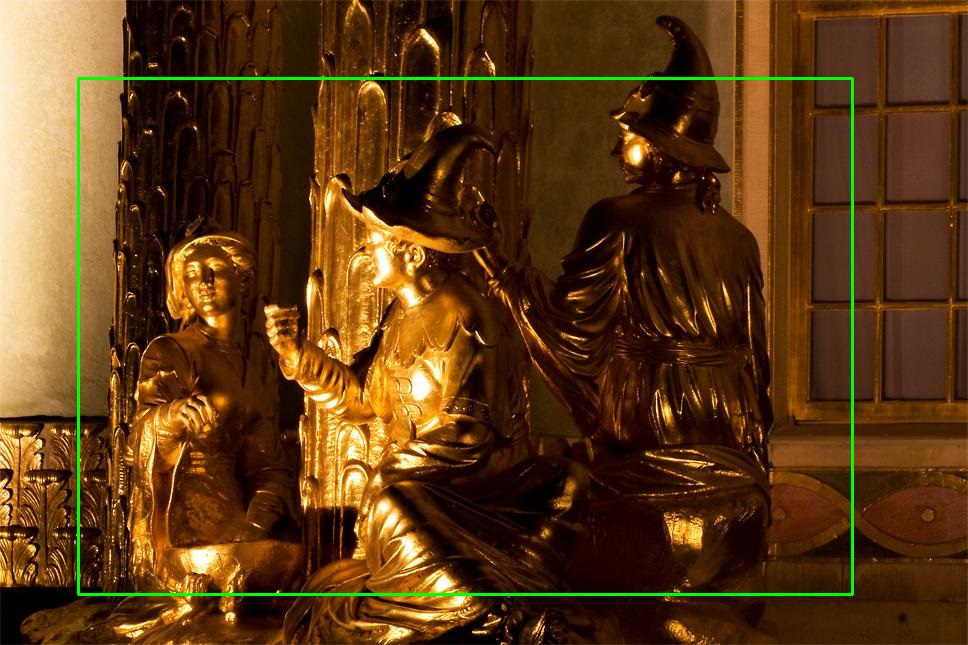}&
\includegraphics[width=3.5cm]{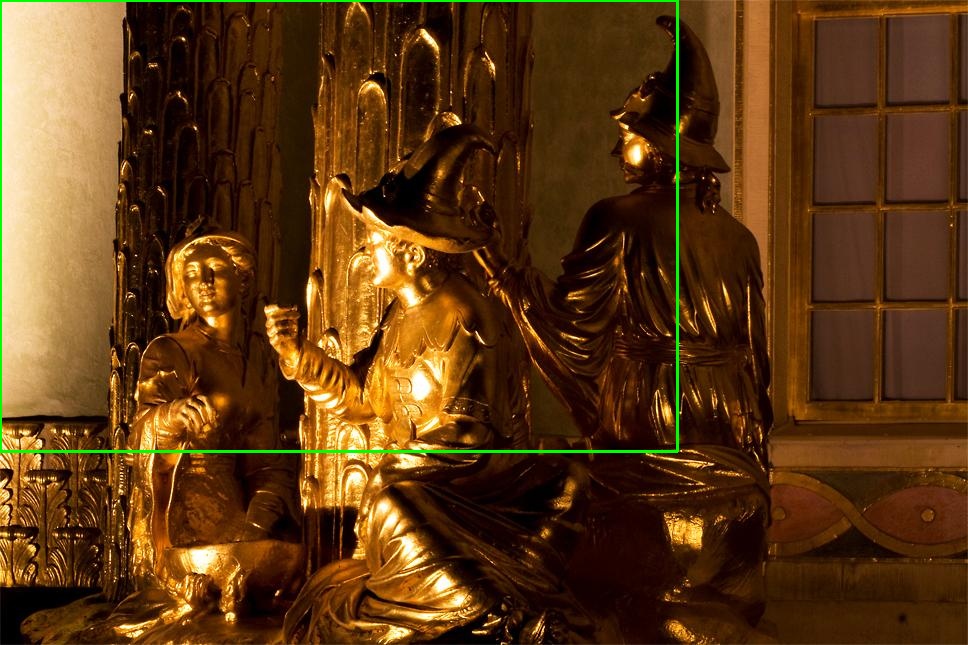}&
\includegraphics[width=3.6cm]{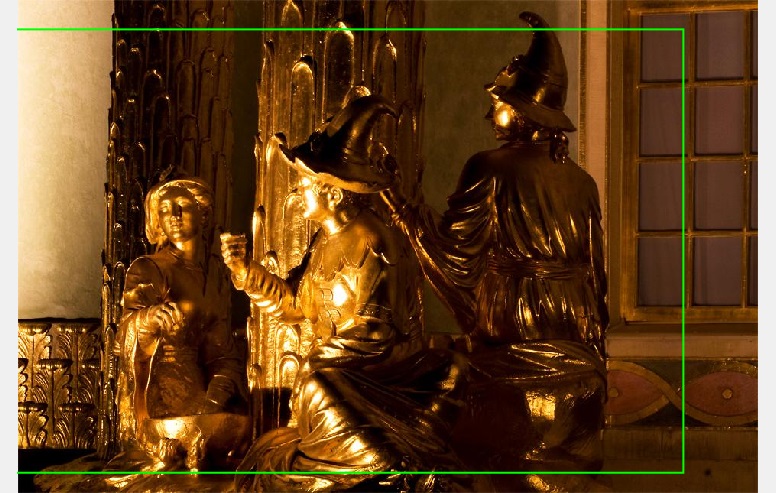}&\\

\includegraphics[width=3.5cm]{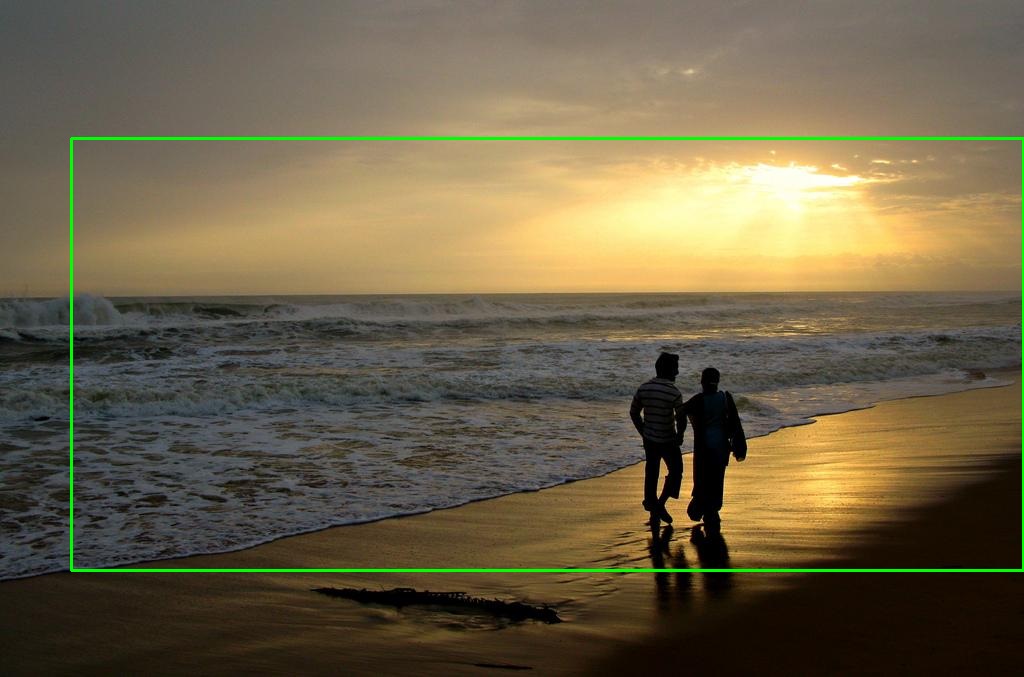}&
\includegraphics[width=3.5cm]{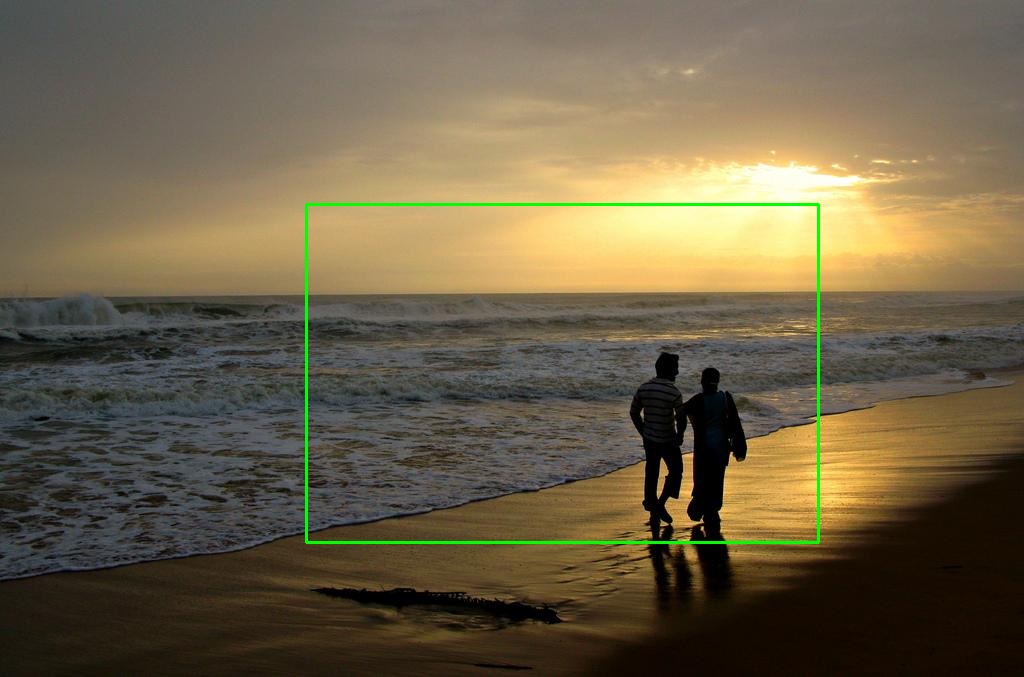}&
\includegraphics[width=3.5cm]{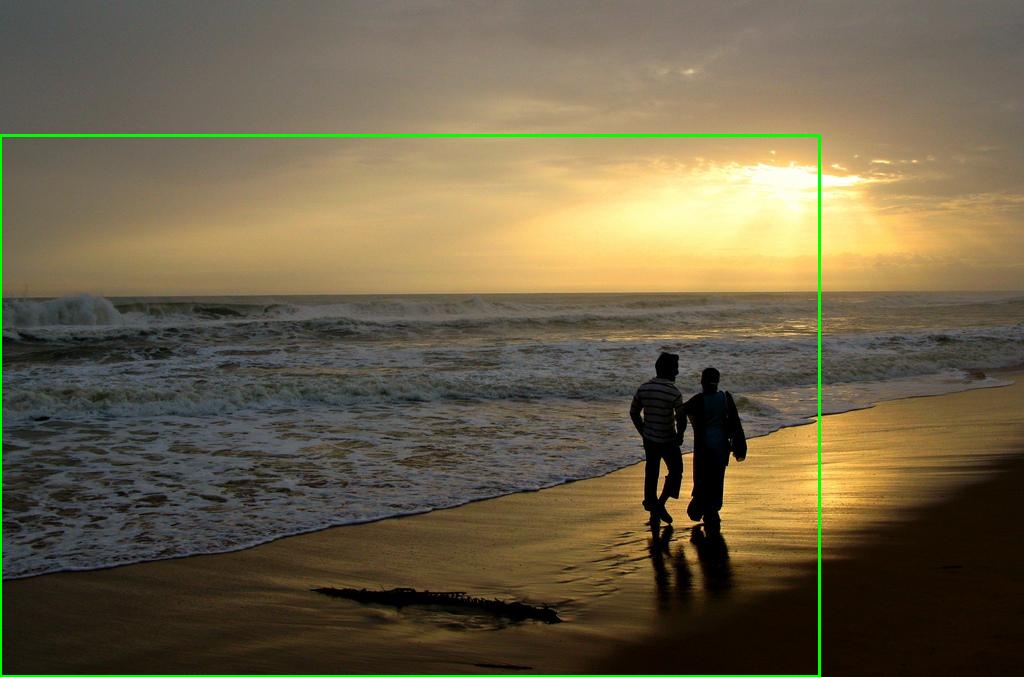}&
\includegraphics[width=3.5cm]{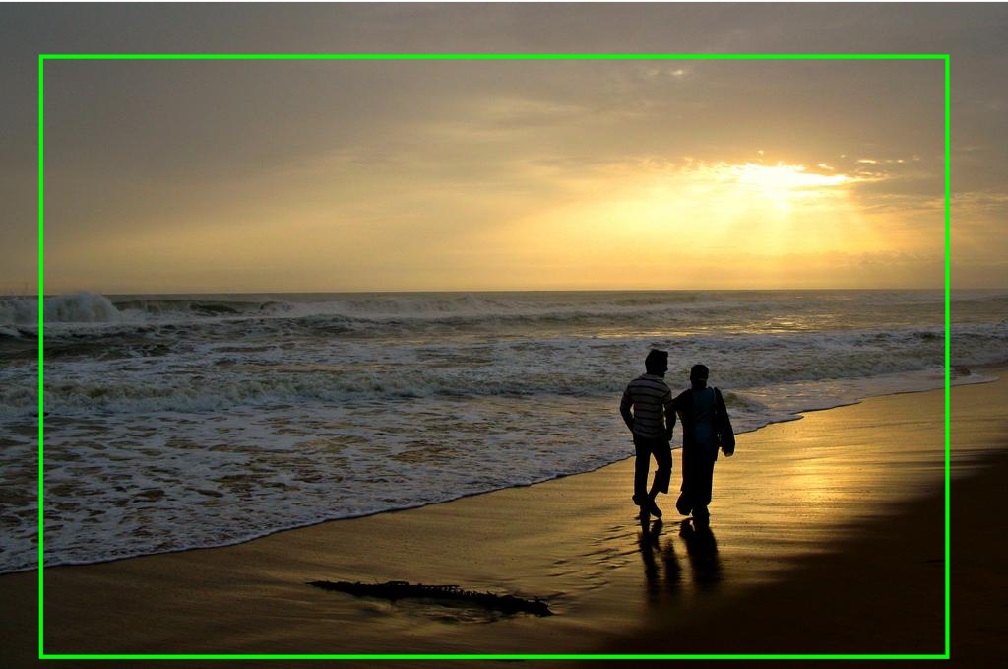}&\\

\includegraphics[width=3.5cm]{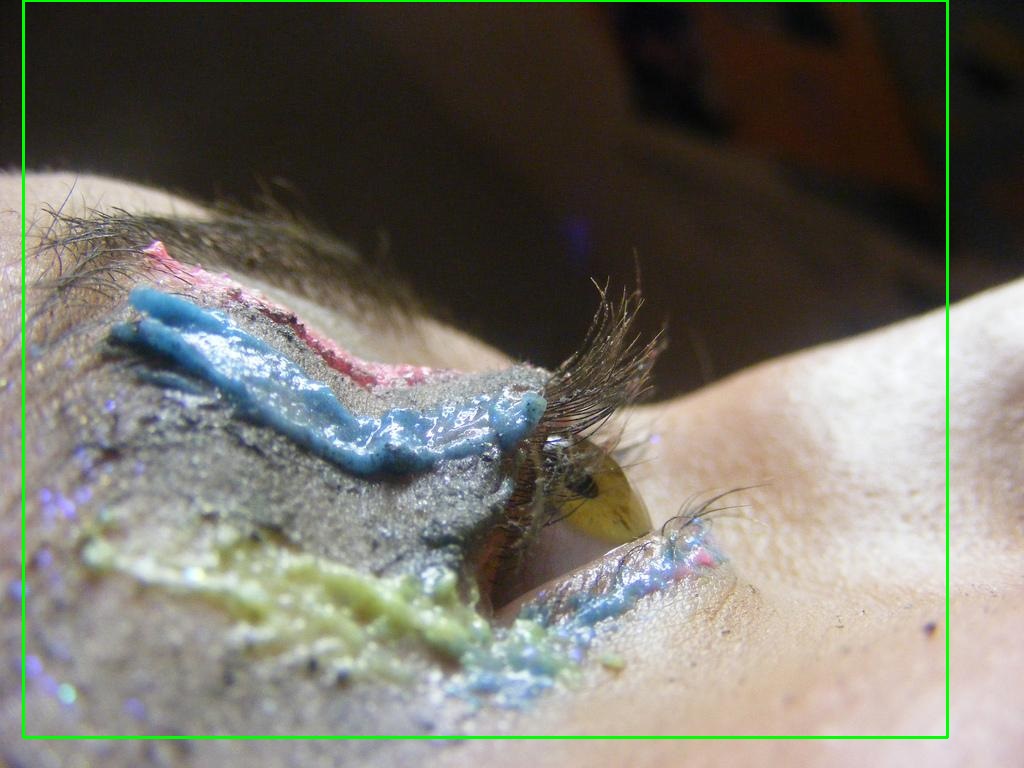}&
\includegraphics[width=3.5cm]{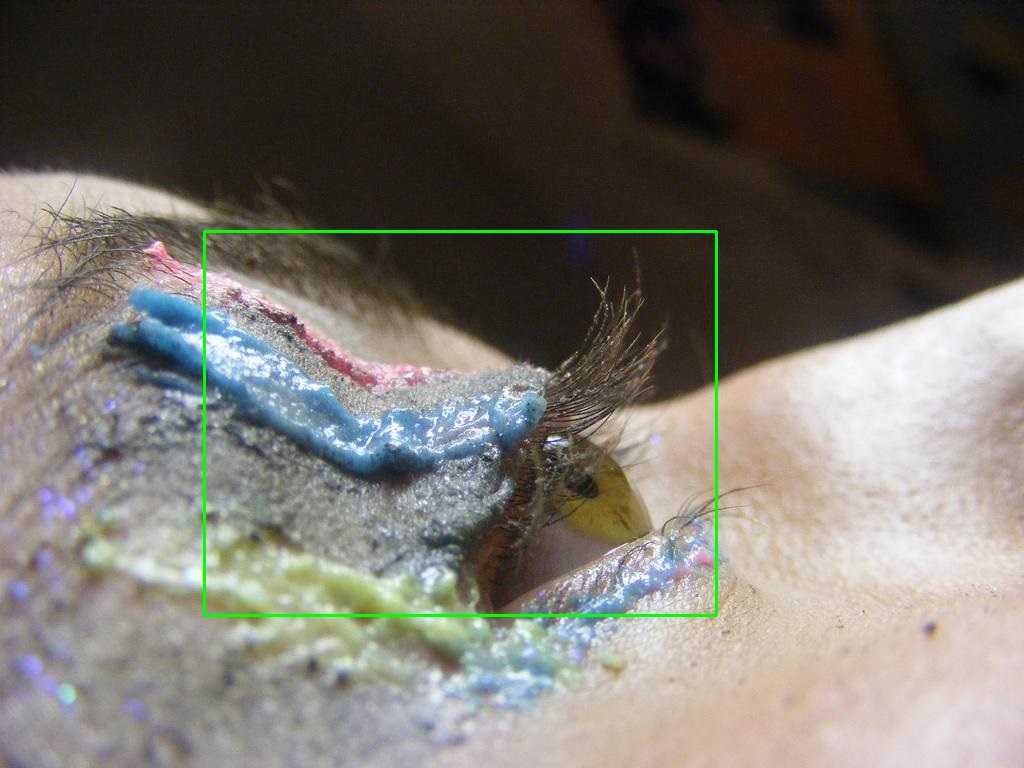}&
\includegraphics[width=3.5cm]{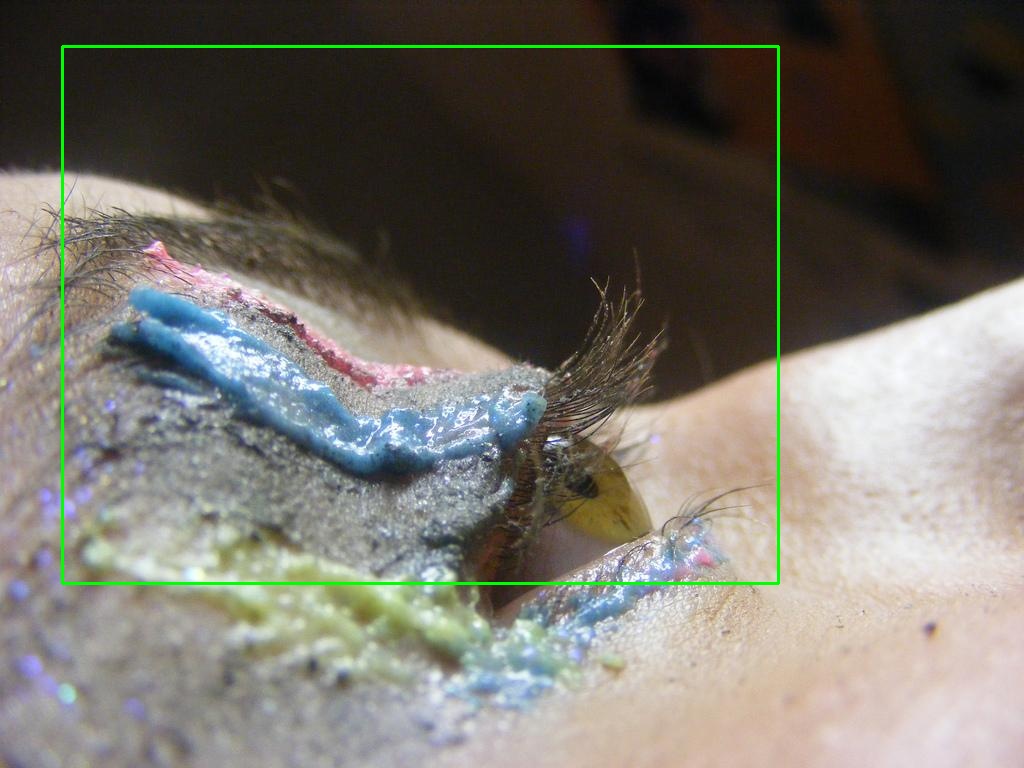}&
\includegraphics[width=3.6cm]{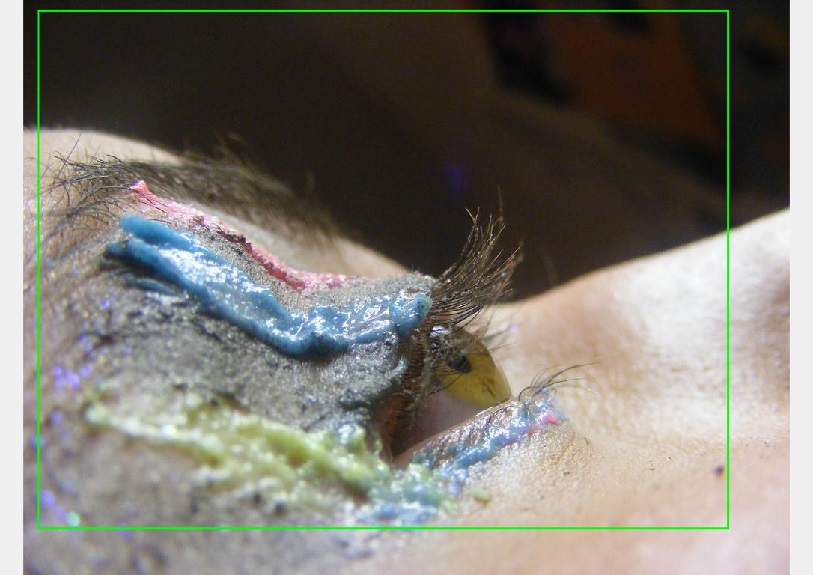}&\\
(a) Ground Truth & (b) \textbf{eDN}(\textbf{MaxDiff}) & (c) \textbf{SVM}+$\mathbf{DeCAF}_7$ & (d) \textbf{Our}+$\mathbf{DeCAF}_7$ \\
\end{tabular}
\caption{Example image cropping results. The optimal crop windows determined by various baselines are drawn as green rectangles.}
\label{fig:baseline_comparison}
\end{figure*}

To summarize, the findings of this study lead to the most important insight of this work: \textit{ranking pairwise views is crucial for image cropping}. 
To the best of our knowledge, all existing methods attempted to tackle this problem by visual saliency detection or learning an aesthetics-aware model from \emph{distinct} images.
However, according to our experimental study, these approaches do not necessarily perform well in differentiating pairwise views with substantial overlaps, which is crucial for image cropping.
\figname \ref{fig:baseline_comparison} demonstrates several examples of comparing the ground truth and the best crop windows determined by various methods.
To maximize the performance of machine learned image rankers, 
two possible directions can be considered: 1) adopting more effective feature representations learned from pairwise ranking of image subviews; 2) developing effective crop selection method to determine potentially good candidate windows for image ranking.

\section{Conclusions}

In this paper, we presented a new dataset which aims to provide a benchmarking platform for photo cropping and view finding algorithms.
With carefully designed data collection pipeline, we were able to collect high quality annotations.
One significant difference between our dataset and other databases is the introduction of pairwise view ranking annotations.
Inspired by the procedure of iteratively comparing in manual image cropping, we argue that learning-to-rank approaches possess great potential in this problem domain, which have been overlooked by most previous researchers.
We conducted extensive study on evaluating the performances of traditional image cropping techniques and several machine learned image rankers.
The experimental results showed that image rankers trained on pairwise view ranking annotations outperform the traditional methods.

\section*{Acknowledgement}

\noindent 
This work was supported in part by Ministry of Science and Technology, National Taiwan University and Intel Corporation under Grants MOST-105-2633-E-002-001 and NTU-ICRP-105R104045.
T.-W.~Huang and H.-T.~Chen are partially supported by MOST 103-2221-E-007-045-MY3.

{\small
\bibliographystyle{ieee}
\bibliography{egbib}
}

\end{document}